\documentclass[conference]{IEEEtran}
\IEEEoverridecommandlockouts
\usepackage{cite}
\usepackage{amsmath,amssymb,amsfonts}
\usepackage{graphicx}
\usepackage{textcomp}
\usepackage{xcolor}
\usepackage{algorithm}
\usepackage{algorithmic}

\usepackage{subcaption}
\usepackage[colorlinks,urlcolor=blue,linkcolor=blue,citecolor=blue]{hyperref}
\usepackage{amsmath,amssymb}
\usepackage{bbm}
\usepackage{verbatim}

\def\BibTeX{{\rm B\kern-.05em{\sc i\kern-.025em b}\kern-.08em
    T\kern-.1667em\lower.7ex\hbox{E}\kern-.125emX}}
\begin{document}

\title{Semi-Targeted Model Poisoning Attack on Federated Learning via Backward Error Analysis}

\author{
    \IEEEauthorblockN{Yuwei Sun\IEEEauthorrefmark{1}\,\IEEEauthorrefmark{2}, Hideya Ochiai\IEEEauthorrefmark{1}, Jun Sakuma\IEEEauthorrefmark{2}\,\IEEEauthorrefmark{3}}
    \IEEEauthorblockA{\IEEEauthorrefmark{1}The University of Tokyo
    \\}
    \IEEEauthorblockA{\IEEEauthorrefmark{2}RIKEN AIP
    \\}
     \IEEEauthorblockA{\IEEEauthorrefmark{3}University of Tsukuba
    \\}
    {ywsun@g.ecc.u-tokyo.ac.jp, ochiai@elab.ic.i.u-tokyo.ac.jp, jun@cs.tsukuba.ac.jp}
}

\maketitle

\begin{abstract}
Model poisoning attacks on federated learning intrude in the entire system via compromising an edge model, resulting in malfunctioning of machine learning models. Such compromised models are tampered with to perform adversary-desired behaviors. In particular, we considered a semi-targeted situation where the source class is predetermined however the target class is not. The goal is to cause the global classifier to misclassify data of the source class. Though approaches such as label flipping have been adopted to inject poisoned parameters into federated learning, it has been shown that their performances are usually class-sensitive varying with different target classes applied. Typically, an attack can become less effective when shifting to a different target class. To overcome this challenge, we propose the Attacking Distance-aware Attack (ADA) to enhance a poisoning attack by finding the optimized target class in the feature space. Moreover, we studied a more challenging situation where an adversary had limited prior knowledge about a client's data. To tackle this problem, ADA deduces pair-wise distances between different classes in the latent feature space from shared model parameters based on the backward error analysis. We performed extensive empirical evaluations on ADA by varying the factor of attacking frequency in three different image classification tasks. As a result, ADA succeeded in increasing the attack performance by 1.8 times in the most challenging case with an attacking frequency of 0.01.
\end{abstract}


\section{Introduction}
Data privacy has become a major concern attracting attention from many walks of life. In particular, the growing public awareness of data privacy and legal restrictions such as the General Data Protection Regulation (GDPR) \cite{gdpr} have rendered classical centralized processing of sensitive data in machine learning (ML) more and more difficult. Due to concerns about critical data breaches in a centralized ML pipeline, decentralized solutions such as federated learning (FL) \cite{fl} have been leveraged to facilitate distributed model training and performance improvement via trained model sharing and aggregation, without disclosing training data of clients.

\begin{figure}[t]
   \begin{subfigure}{0.49\textwidth}
     \centering
    \includegraphics[width=0.9\linewidth]{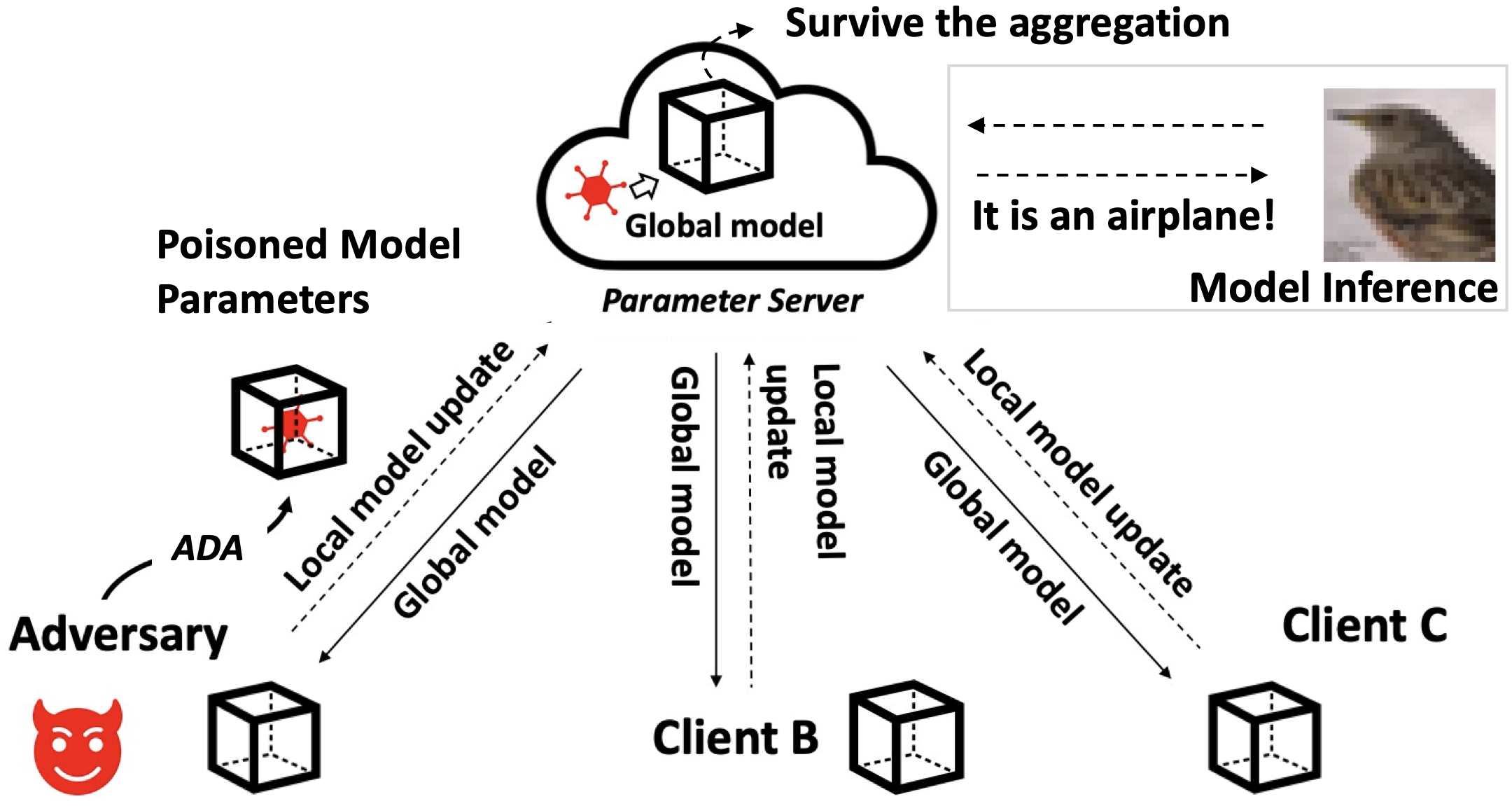}
    \caption{Model poisoning in FL.}
       \label{intro}
   \end{subfigure}\hfill
   \begin{subfigure}{0.49\textwidth}
     \centering
     \includegraphics[width=0.85\linewidth]{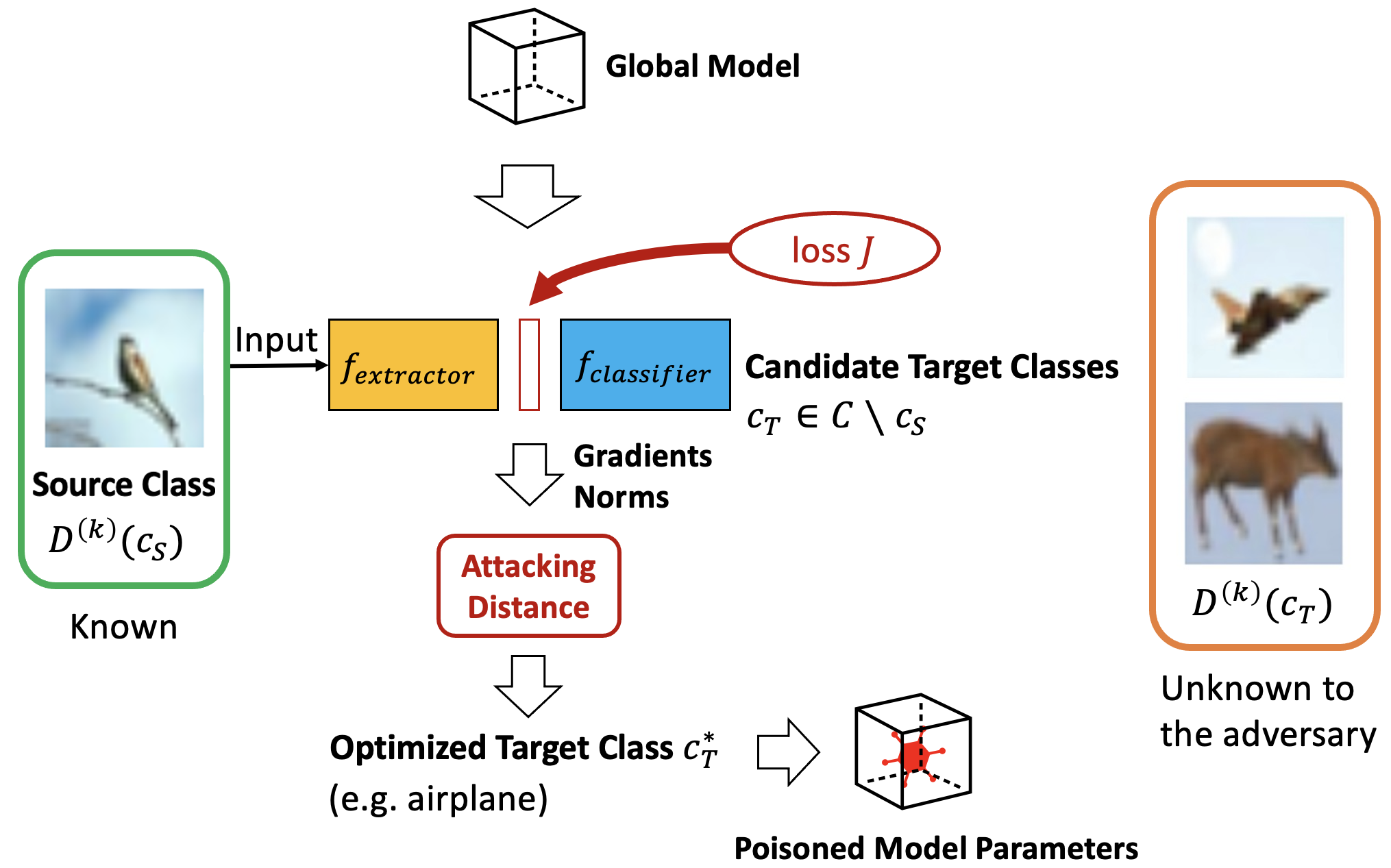}
     \caption{Semi-targeted attack mounted by the adversary.}
        \label{intro2}
   \end{subfigure}
   \caption{Semi-targeted Attacking Distance-aware Attack (ADA).}
\end{figure}

Despite its broad application in lots of fields such as medical image classification, financial data analysis, cybersecurity, and so on, FL has been shown venerable to adversarial attacks \cite{labelflipping, WangSRVASLP20, 45}. A compromised edge client can inject malicious parameters into the entire FL systems, causing malfunctioning of ML models and extending influence to other clients in the systems (Fig.\ref{intro}). Moreover, these attacks are usually either untargeted or targeted \cite{survey}. The goal of the untargeted attacks is to degrade a classifier's performance in general. In contrast, the targeted attack aims to compromise the classifier such that it misclassifies samples of a specified class into an adversary's desired class.

In this study, we focus on a novel type of poisoning attack, where the motivation of the attacker is avoiding being recognized as a specific class. This type of adversary can arise in various real-world situations. For example, an attacker who sends unauthorized advertising mails aims to have that mail recognized as belonging to some benign class other than spam. In a facial recognition system, a person on a blacklist aims to be recognized by someone else not on the blacklist. Note that these attackers are not motivated to be recognized as a particular class (as the targeted attack aims to), but rather to be unrecognized as a particular class. 

We formulate this type of attack as a semi-targeted attack. In the semi-targeted attack, we suppose the adversary is assigned a class (source class), and the aim of the adversary is to poison the classifier so that samples with the source class are recognized as a class other than the source class. Different from the targeted attack, the adversary can freely choose a target class so that the attack performance is maximized in the semi-targeted attack. Due to this freedom of attack, the risk of the semi-targeted attack outweighs the risk of the targeted attack always.

Depending on the assigned source class, the attack's successability can be different. In other words, the attack is typically class-sensitive regarding generality, attack performance varying from case to case with various classes considered \cite{BagdasaryanVHES20}. The challenge in the semi-targeted attack is how to find the target class optimized for the assigned source class resulting in the best attack performance.

We proposed two approaches to find the optimized target class to mount an attack in the full knowledge and partial knowledge settings respectively. Our goal is to investigate to what extent the attack performance can be increased in the semi-targeted setting regarding different approaches to choosing the target classes. 

Overall, our main contributions are as follow:

1) We introduced a new type of poisoning attack called the semi-targeted attack, provided a detailed formulation, and studied the risk of this type of attack in a federated setting.

2) We proposed a semi-targeted model positioning attack called the ADA to compromise the classifier by carefully choosing the target class (Section \ref{fullknowledge}). Based on a backward error analysis of the global model, the update gradient's scale for producing a poisoned model was computed to measure the expected attack performance when choosing a certain class (Fig.\ref{intro2}). In this regard, we demonstrated the Fast LAyer gradient MEthod (FLAME) in the partial knowledge setting (Section \ref{partialknowledge}).

3) We performed an extensive study of the proposed methods by varying the factor of attacking frequency and comparing its performance with other methods against the metrics of adversary task accuracy (ATA) and main task accuracy (MTA) using MNIST, Fashion-MNIST, and CIFAR-10. The empirical results showed that ADA worked effectively in both full knowledge (white-box) and partial knowledge (black-box) settings. ADA succeeded in increasing the attack performance by 1.8 times when the attacking frequency is as low as 0.01 (Section \ref{experiments}).

The remainder of this paper is structured as follows. Section 2 reviews the most recent work of poisoning attacks on FL. Section 3 presents essential definitions and assumptions. Section 4 demonstrates the technical underpinnings of the proposed method. Section 5 presents extensive empirical evaluations. Section 6 concludes the paper and gives out future directions.

\section{Related Work}
Federated Learning (FL) is encountered with various types of adversarial attacks. Notably, as for poisoning attacks, an adversary usually compromises an FL system by manipulating either training data \cite{BagdasaryanVHES20,29,30,WangSRVASLP20} or local training process \cite{WangSRVASLP20, 32} of an edge client. We briefly review related research efforts of poisoning attacks on FL in the following.

The goal of an adversary in a targeted poisoning attack on FL is to produce a poisoned local model update, such that after model aggregation particular inputs will induce misclassification of the global model \cite{datapoison1, datapoison2}. In particular, label flipping is one type of poisoning attack, where a set of data's labels are randomly flipped to a different class for malicious model training. For example, a semantic backdoor flips the labels of images with specific natural features to cause the misclassification with these features as triggers \cite{WangSRVASLP20}. 

Moreover, a malicious model update based on the label flipping usually yields a large norm of model weights compared to a benign update. To prevent the attacks from being easily detected by a norm-based defense algorithm in FL, Bagdasaryan et al.\cite{BagdasaryanVHES20} presented a train-and-scale technique that scaled the norm of a malicious update to the bound of the detecting algorithm. Whereas, this strategy could also result in degraded attack performance due to diminished poisoned model weights. Furthermore, the model replacement attack \cite{BagdasaryanVHES20} aimed to completely replace the global model with a malicious adversary's model. Notably, due to the converged global model resulting in a relatively small benign local update, an adversary-controlled client could upload a crafted malicious update that eventually replaced the global model after the aggregation.

Though the aforementioned targeted attacks aim to compromise an FL system, their attack performance typically depends on the assigned source class. There does not exist work on semi-targeted attacks that have a fixed source class and an adjustable target class. A semi-targeted attack tackles the problem that a poorly chosen target class for the assigned source class degrades the poisoning attack's performance after the model aggregation in FL and even results in more easily discovered poisoning parameters. To the best of our knowledge, our work is the first to study the semi-targeted poisoning attack with the consideration of generality in FL. In this regard, Shafahi et al.\cite{shafahi2018poison} presented a feature-collision method aimed to generate similar-looking instances based on the source class such that their hidden features were close to the target class in a centralized setting. To mount the attack, the adversary needed prior knowledge about both the data distributions of the source class and the target class, which is different from our method with the federated setting where neither the data modification nor the prior knowledge about the target class's data distribution is required. Furthermore, we studied poisoning attacks in FL with a more realistic setting regarding the attacking frequency, where the adversary had only a small possibility of participating in the training every round.

\section{Preliminaries}
\subsection{Classification Task}
We specifically consider supervised learning with $C$ categories in the entire dataset $D$. Let $x \in \mathbb{R}^V$ be a sample and $y \in \{1,2,...,C\} = Y$ a label. $D$ consists of a collection of $N$ samples as $D=\{(x_i, y_i)\}_{i=1}^N$. Suppose that $f$ denotes a neural network classifier taking an input $x_i$ and outputting a $C$-dimensional probability vector where the $j$th element of the output vector represents the probability that $x_i$ is recognized as class $j$. Given $f(x)$, the prediction is given by $\hat{y} = \mbox{arg}\max_j f(x)_j$ where $f(x)_j$ denotes the $j$th element of $f(x)$. The training of the neural network is attained by minimizing the following loss function with respect to the model parameter $\theta$ 
\begin{equation}
J(\theta, D)= \frac{1}{N}\sum_{i=1}^N \ell(y_i, f(x_i; \theta)).
\end{equation}

\subsection{Federated Learning}
We assume there are $K$ clients. The $k$th client has its own dataset $D^{(k)}:=\{(x_i, y_i)\}_{i=1}^{N^{(k)}}$ where $N^{(k)}$ is the sample size of dataset $D^{(k)}$. Here, $\cup_{i=1}^K  D^{(k)} = D$ and $N=\sum_{k=1}^{K} N^{(k)}$. We also suppose that each client cannot share data mainly due to data confidentiality.

FL is a framework to obtain a global model that is trained over the entire data without disclosing distributed training samples. The process of FL proceeds by iterating the following steps: (1) the parameter server (PS), that controls the entire process of FL, initializes the global model $G_0$ and delivers it to all clients, (2) each client updates the model using local data $D^{(k)}$, and sends back its model update $L_{t+1}^{(k)}-G_t$ to PS, (3) then, PS aggregates all local model updates, updates the global model, and sends the global model to all clients. Here, we suppose that the neural networks of clients' local models and the global model share the same architecture in FL.

Moreover, typically, to reduce the waiting time for all clients completing local model training, PS randomly selects a subset of clients $K_t$ every round to updating the global model based on local updates sent from these clients. We introduce a typical FL algorithm, FedAvg\cite{17}, where the averaging of all local updates is employed to update the global model. The detailed FedAvg algorithm is shown in Algorithm \ref{fedavg}.

\begin{algorithm}[tb]
\caption{FedAvg}
\label{fedavg}
\begin{algorithmic}[1]
\STATE initialization of $G_0$ at the server side
\FOR {each round $t = 0, 1, 2, . . .$}
    \STATE PS randomly selects a subset of clients $K_t$ from all clients $K$
    \STATE PS sends the current global model $G_t$ to $K_t$ 
    \FOR {each client $k = 1, 2, . . ., K_t$}
        \STATE $L_{t+1}^{(k)}$ = FedLearnLocal($G_t, L_{t}^{(k)}, D^{(k)}$)
    \ENDFOR
    \STATE $G_{t+1} = G_t +\sum_{k\in K_t}\frac{N^{(k)}}{\sum_{k\in K_t} N^{(k)}}(L_{t+1}^{(k)} - G_t)$
\ENDFOR
\STATE
\STATE \textbf{function} FedLearnLocal($G_t, L_{t}^{(k)}, D^{(k)}$)
    \STATE  $L_t^{(k)}\leftarrow G_t$
    \FOR {each epoch $e \in E$}
        \FOR {each batch $b \subset D^{(k)}$}
            \STATE $L_{t+1}^{(k)} \leftarrow L_{t}^{(k)}-\eta\cdot\nabla J (L_{t}^{(k)}, b)$
        \ENDFOR
    \ENDFOR
    \RETURN $L_{t+1}^{(k)}$
\end{algorithmic}
\end{algorithm}

\subsection{Poisoning Attacks on Federated Learning}
Poisoning attacks have been well explored in the centralized learning setting. Considering a supervised classification model $f$ for $C$ categories, the goal of the adversary is to either degrade the classifier's performance in general (untargeted) or trigger the misclassification of the classifier on a specific class (targeted). 

In the targeted poisoning attacks, we call the class of samples used for attacks the source class $c_S$ and the final class which a sample is modified to, the target class $c_T$. Then, given a classifier $f$ and a sample $(x, y)$, the goal of the targeted poisoning attacks is to corrupt $f$ so that 
\begin{equation}
\mbox{arg}\max_k f(x)_k = 
\begin{cases}
c_T \mbox{ if } y=c_S, \\
y \mbox{ Otherwise }.
\end{cases}
\label{eq:poisoning}
\end{equation}

Let $D_{\mbox{val}}$ be a validation dataset, $D_{\mbox{val}}(c_S) $ be the set of samples in $D_{\mbox{val}}$ with label $c_S$, and $D^{c}_{\mbox{val}}(c_S) = D_{\mbox{val}} \setminus D_{\mbox{val}}(c_S)$.
To measure the performance, the main task accuracy (MTA) evaluates the success rate of the classifier for the main task, the classification of samples that are not from the source class. In particular, MTA of classifier $f$ poisoned with source label $c_S$ is defined as the validation accuracy for $D^{c}_{\mbox{val}}(c_S)$:
\begin{equation}
\mbox{MTA}_f(c_S) = \frac{  \sum_{ (x,y) \in D^{c}_{\mbox{val}}(c_S)}   \mathbbm{1}\{ \mbox{arg max}_{k}f(x;G)_k  = y   \}    }{| D^{c}_{\mbox{val}}(c_S) |}.
\end{equation}

The target-specified adversary task accuracy (ts-ATA) evaluates the success rate of an attack. The attack succeeds if the poisoned model outputs the desired target class for a sample in the source class, otherwise the attack fails. ts-ATA of classifier $f$ poisoned with source label $c_S$ is defined as the validation accuracy for $D_{\mbox{val}}(c_S) $:
\begin{equation}
    \mbox{ts-ATA}(c_S, c_T) = \frac{   \sum_{ x\in D_{\mbox{val}}(c_S)}   \mathbbm{1}\{ \mbox{arg max}_{k}f(x;G)_k  = c_T   \}    }{  |D_{\mbox{val}}(c_S)|     }.
\end{equation}

We introduce two building blocks, label flipping \cite{labelflipping} and gradient scale adjustment by Train-and-scale technique \cite{BagdasaryanVHES20}, needed to introduce to our proposed method. 

\subsubsection{Label Flipping Attack}

In this attack, samples from the source class are relabeled with the target class.
For samples $(x,y) \in D(c_S)$, poisoning by label flipping is attained by replacing the label by $c_T$ as $(x, c_T)$.
By training the classifier with the samples containing the poisoned samples, we can expect that the poisoning attack in eq. \ref{eq:poisoning} is achieved.
The adversary usually adopts a mix of poisoned samples and legitimate samples. 
In our experiments, we perform label flipping to all samples in $D(c_S)$, and a subset of other samples are randomly selected so that the poisoned samples account for $\alpha$ percent of the total training samples of the adversary, where $\alpha$ is called injection rate. 
Let $D^{adv}$ denote the mixture of poisoned samples and benign samples. This approach aims to mount a poisoning attack without sacrificing too much on the accuracy of other non-target classes.

In an FL setting, if a compromised client is selected by PS for local model training, it will download the latest global model to replace its local model, train a local model $L^{adv}$ with the poisoned samples, and corrupt the global model by iteratively submitting malicious updates to the PS. In addition, multiple compromised clients could exist to mount poisoning attacks against the global model.

\subsubsection{Gradient Scale Adjustment by "Train and Scale"} 

The aforementioned malicious model updates are prone to producing large norms compared with other legitimate updates. From the PS's perspective, an efficient way to detect malicious updates will be to drop the updates whose norm is above a certain threshold $Q$, i.e., $\|L_t^{(k)}-G_t\| > Q$ here $\|\cdot \|$ is a prescribed norm. This defense technique is referred to as norm difference clipping (NDC)\cite{abs-1911-07963}, where the norm of a local update can be defined arbitrarily so that it reflects the difference of updates.

On the other hand, from the perspective of the attacker side, defense strategies that consider the magnitudes of local updates, such as NDC, can be readily bypassed by the "Train and Scale" approach. 
Notably, to achieve this goal, the adversary modifies the scale of the model update $\|L^{adv}_{t+1}-G_t\|$ so that it is upper bounded by $Q$.
Let a scaling factor $\gamma$ defined by
\begin{equation} 
\gamma=\frac{Q}{\|L^{adv}_{t+1}-G_t\|}.
\end{equation}
Then, the malicious client submits $\gamma \|L^{adv}_{t+1}-G_t\|$ as model update instead of $\|L^{adv}_{t+1}-G_t\|$, which is called "Train and Scale" technique.

One limitation of this approach is that $Q$ is usually unknown to the adversary, however, the adversary can estimate $Q$ in the following simple strategy. 
We can expect that the PS's $Q$ is set so that regular updates are not rejected with high probability.
With this consideration, the adversary can first adopt several compromised clients to perform legitimate training with the latest global model shared by the PS, and the average norm of the collected legitimate updates can be employed as a lower bound of $Q$.

\section{Attacking Distance-aware Attack}
In this section, we first introduce the motivation for introducing the new type of poisoning attack called the semi-targeted attack. We then discuss the Attacking Distance-aware Attack (ADA) in detail. Finally, we demonstrate the Fast LAyer gradient MEthod (FLAME) to mount the semi-targeted attack in the partial knowledge setting.

\subsection{Motivation}
In a scenario of multi-class classification tasks, an adversary aims to compromise the system such that instances from a specific class $c_S$ will be misclassified. Compared to a targeted attack with a fixed target class $c_T$ for the given source class $c_S$, the semi-targeted attack that does not have a specified target class could provide more relaxation which could increase the risk of an ML system. For example, in the real world, a self-driving car that recognizes a stop sign could be compromised such that the prediction of the stop sign will be wrong. The incorrectly predicted class can be the speed limit sign, the billboard, and so on. Moreover, a spam filter aims to identify the category of an email that can be poisoned such that the risk of a certain type of spam will bypass the filter. In this case, the target class can be sports, politics, and so forth.        

Regarding the semi-targeted attack, the adversary's performance can vary from case to case with different target classes $c_T$ considered. Its robustness to different attacking cases is barely guaranteed. An attack based on a less effective target class would require a longer convergence time to achieve the same attack performance. Furthermore, a longer convergence time allows a defense strategy in the PS to more easily discover the attack. In addition, due to a small part of clients being selected every round, outnumbering legitimate updates can overwrite a less effective malicious update. 
 
Intuitively, if the adversary can choose the most effective target class, a poisoning attack will be greatly enhanced. To reveal the risk of the semi-targeted attack, we propose the Attacking Distance-aware Attack (ADA), an enhanced model poisoning attack on FL. Notably, ADA measures the distances in the latent feature space between different classes of a classifier and finds the optimized target class $c_T$ given a source class $c_S$ to mount the attack. Based on the prior knowledge of the adversary about a client's data distribution in FL, we studied two different settings of ADA, i.e., attacking with full knowledge and attacking with partial knowledge.

\subsection{Semi-Targeted Attack}
The semi-targeted attack in FL refers to a model poisoning attack with a fixed source class $c_S$ and various possible target classes $\{Y\setminus c_S\}$. the goal of the semi-targeted poisoning attacks is to corrupt $f$ so that
\begin{equation}
\mbox{arg}\max_k f(x)_k = 
\begin{cases}
c^* \mbox{ if } y=c_S, \\
y \mbox{ o.w. }.
\end{cases} 
\end{equation}
where 
\begin{equation}
c^* = \mbox{argmax}_c \mbox{ts-ATA}(c_S, c).
\end{equation}

In particular, for any given $c_S$, $\mbox{ts-ATA}(c_S, c_T)$ depends on the choose of $c_T$. In the semi-targeted attack, the attacker could choose any arbitrary $c_T$ thus the attack performance could be increased. 
In this regard, the max target-non-specified adversary task accuracy (max-ATA) of classifier $f$ poisoned with source label $c_S$ is defined as the validation accuracy for $D(c_S) $:
\begin{equation}
    \mbox{max-ATA}(c_S) = \max_{c \in C }  \mbox{ts-ATA}(c_S,c).
\end{equation}

Note that,
\begin{align*}
    \mbox{max-ATA}(c_S)  \geq \mbox{ts-ATA}(c_S, c_T)
 \end{align*}  
holds for any $c_S$ and $c_T$, which means that the semi-targeted poisoning attack is always more powerful than targeted poisoning due to the more relaxed constraint.
 
In addition, for simplicity, we assume the number of source classes $c_S$ specified by the adversary is one, however, this semi-targeted attack can also be extended to scenarios with multiple source classes where the optimized target class is computed for each source class.   

\subsection{Attack Method}

In reality, finding $c^*$ is not tractable. In this subsection, we discuss how the adversary finds a target class that is close to $c^*$ depending on the adversary's knowledge of the training samples of the shared global model.

\subsubsection{Attacking with Full Knowledge}
\label{fullknowledge}

In the full knowledge setting, we suppose the adversary is allowed to access samples that are drawn from the underlying training sample distribution in an independent and identically distributed (I. I. D.) manner. 

To choose the class that is supposed to give larger ts-ATA, the adversary leverages the attacking distance (AD) defined as follows.
Let $\phi: \mathbb{R}^V \rightarrow \mathbb{R}^U$ be the feature extractor of the shared global model $G_t$.
Then the adversary extract feature vectors $\phi(x)$ of local training samples $x \in D^{adv}$. 

Let $\mu_c$ be the mean of feature vectors in class $c$ where
\begin{equation}
\mu_c = \frac{1}{|D^{adv}(c)|} \sum_{x \in D^{adv}(c)} \phi(x).
\end{equation}
Then, the attacking distance between class $c$ and $c'$ is defined by
\begin{equation}
\label{eq:ad}
\mbox{AD}(c,c') = \|  \mu_c - \mu_{c'}  \|_2
\end{equation}
where $\|  \cdot \|_2$ denotes the $\ell_2$ norm.

\begin{figure}[t]
\centering
\includegraphics[width=0.92\linewidth]{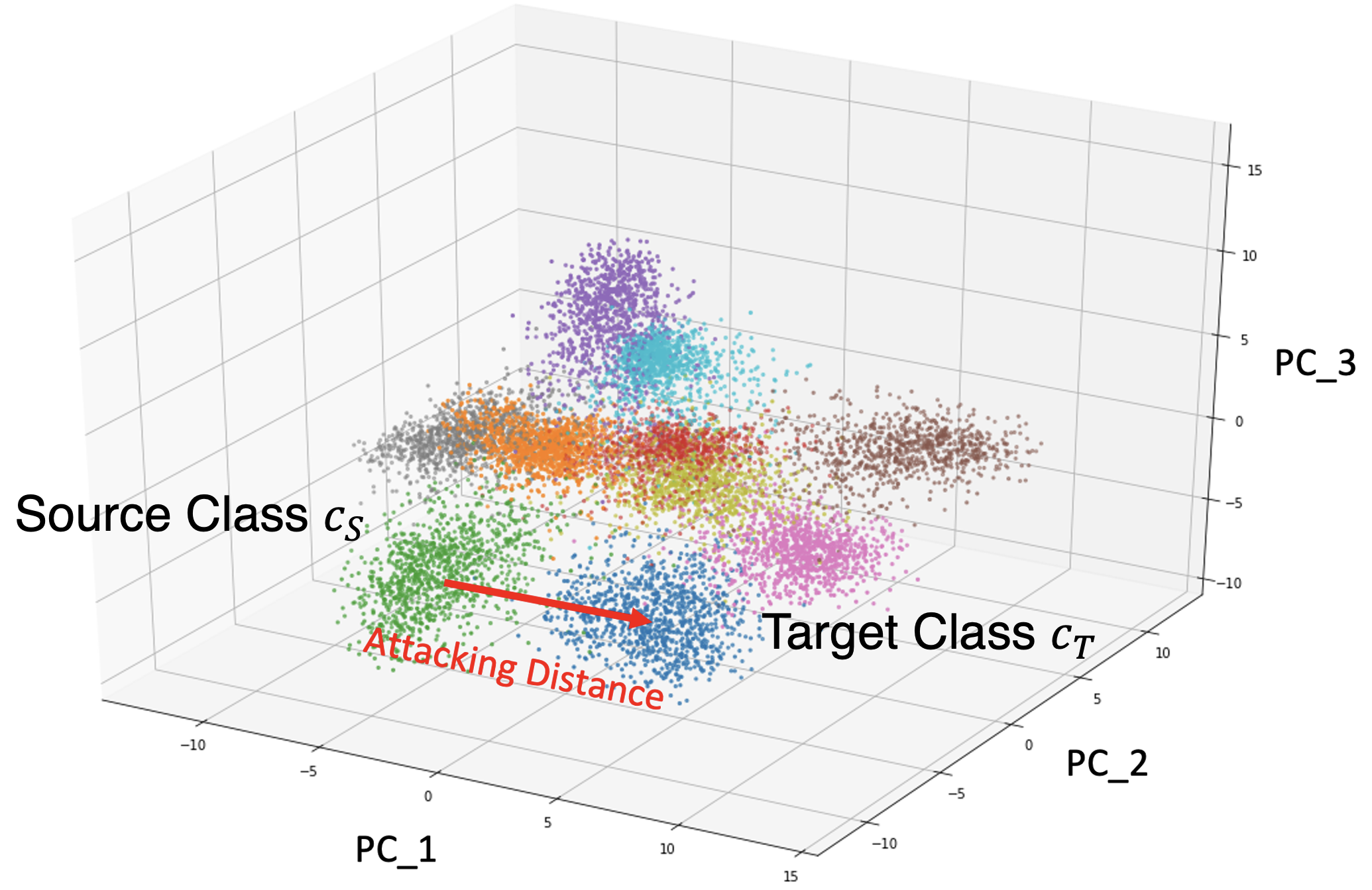}
\caption{Distribution visualization of the extracted feature vectors $\phi$ using the PCA.}
\label{pca}
\end{figure}

The distribution visualization of the extracted feature vectors $\phi$ using the principal component analysis (PCA) for measuring the AD is shown in Fig.\ref{pca}.  

The idea of the attack in this setting is that if the AD between the source class $c_S$ and a target class $c_T$ is small, the scale of the required malicious update in the adversary's local model is small. Then, the attack has a larger chance of surviving the aggregation with other outnumbering legitimate updates, as a result, with more effect. For this reason, with the AD\label{eq:ad}, the target class is determined by the following
\begin{equation}
\label{eq:adafull}
c^*_T = \mbox{argmin}_{c \in C\setminus c_S}\mbox{AD}(c_S,c_T) 
\end{equation}

Furthermore, the adversary performs malicious model training based on the optimized target class $c_T^*$ with an injection rate $\alpha$, where label flipping is applied to all samples in $D(c_S)$ and a subset of other samples are randomly selected. $L_{t+1}^{adv} = L_{t}^{adv}-\eta\cdot\nabla \ell(L_{t}^{adv},D^{adv})$, where $D^{adv}:=\{\{(x_i, c_T^*)\}_{i=1}^{\lfloor \alpha N^{(k)}\rfloor}, \{(x_i, y_i)\}_{i=(\lfloor \alpha N^{(k)}\rfloor+1)}^{N^{(k)}}\}$. Then, the adversary scales the poisoned model update by $\gamma$ to the bound of the norm-based defense and sends the scaled malicious update $\gamma(L^{adv}_{t+1}-G_t)$ to the PS. 

\subsubsection{Attacking with Partial Knowledge}
\label{partialknowledge}

The aforementioned attack is limited to the I.I.D. setting. Whereas, in real life, it is usually difficult and even impossible for the adversary to know the entire feature space in FL. In other words, it is unfeasible to extract the latent feature representations of all classes and measure their AD due to only partial classes' representations being available to the adversary. 

To tackle the case of attacking with partial knowledge, we propose a backward error analysis method called the Fast LAyer gradient MEthod (FLAME) to find a class that is supposed to be $c^*$ without the need for prior knowledge about the entire sample distribution.

Simple patterns such as lines and curves activate neurons in the shallower layer while more complex features in the deeper layer\cite{Goodfellow-et-al-2016}.
For samples in the source class, by arbitrarily assigning a wrong label and computing the loss, FLAME measures the scale of the update required in the higher layer of the applied model. The update scale reveals the distance between the current model's inner state of the latent feature representation and the poisoned model that misclassifies the input as the wrong label.

Instead of measuring the distance between classes based on the output of the higher layer in the full knowledge case, FLAME employs the update scale to estimate the distance between the source class and the assigned class. In particular, if the update scale in the higher layer is large when assigning a specific class, the distance between the latent feature representations of the source class and the assigned class is assumed also large. In contrary, if the ground truth label is assigned, the update scale should be close to zero.  

To realize the intuition described above, we employ the following approach. 
Similar to the case in the full knowledge setting, for each round, a selected compromised client $k$ downloads the latest global model $G_t$ to replace its local model $L_t^{(k)}$. The adversary then inputs the source class samples into the local model and forward-propagate to obtain the output of the model. The loss between the prediction and an assigned label is computed based on cross entropy and back-propagated to compute the update gradients with respect to the last FC layer.
Let $1_c$ denotes the one-hot vector where the $c$th element is $1$ and remaining elements are $0$.
Let $CE(f(x), 1_c)$ represents the cross entropy loss between $y=f(x)$ and $1_c$.
Then, the total loss of samples in $D^{(k)}(c_S)$, the adversary's samples labeled with the source class, when the ground truth label for all samples are set to $c$ is 
\begin{equation}
J(L_t^{(k)}) =\sum_{x \in D^{(k)}(c_S)}  CE(f(x), 1_c).
\end{equation}
With this empirical loss, the target class is determined by the following
\begin{equation}
\label{eq:flame}
c^*_T = \mbox{argmin}_{c \in C\setminus c_S}  \|  \nabla J(L_t^{(k)}) \|_2  
\end{equation}
where the derivative is taken with respect to the weight parameters of the last FC layer in $L_t^{(k)}$.

Finally, the intact ADA algorithm in the partial knowledge setting with the FLAME adopted is shown in Algorithm \ref{ada}. 

\begin{algorithm}[tb]
\caption{ADA in the Partial Knowledge Setting}
\label{ada}
\begin{algorithmic}[1]
\STATE initialization of $G_0$ at the server side
\FOR {each round $t = 0, 1, 2, . . .$}
    \STATE PS randomly selects a subset of clients $K_t$ from all clients $K$
    \STATE PS sends the current global model $G_t$ to $K_t$
    \FOR {each client $k \in K_t$}
    			\IF{$k\in K^{adv}$ and global model converged}
                	\STATE {$L_{t+1}^{(k)}$ = ADA($G_t$)}
                \ELSE
        			\STATE $L_{t+1}^{(k)}$ = FedLearnLocal($G_t$)
                \ENDIF          
    \ENDFOR
    \STATE $G_{t+1} = G_t +\sum_{k\in K_t}\frac{N^{(k)}}{\sum_{k\in K_t} N^{(k)}}(L_{t+1}^{(k)} - G_t)$
\ENDFOR
\STATE
\STATE \textbf{function} ADA($G_t$)
    \STATE $L_t^{(k)}\leftarrow G_t$
    \STATE $J(L_t^{(k)}) =\sum_{x \in D^{(k)}(c_S)}  CE(f(x), 1_c)$
	\STATE $c^*_T = \mbox{argmin}_{c \in C}  \|  \nabla J(L_t^{(k)}) \|_2$, where the derivative is taken with respect to the weight parameters of the last FC layer in $L_t^{(k)}$
    \STATE $D^{adv}:=\{\{(x_i, c_T^*)\}_{i=1}^{\lfloor \alpha N^{(k)}\rfloor}, \{(x_i, y_i)\}_{i=(\lfloor \alpha N^{(k)}\rfloor+1)}^{N^{(k)}}\},$\\
    \FOR {each epoch $e \in E$}
        \FOR {each batch $b\in D^{adv}$}
            \STATE $L_{t+1}^{adv}\leftarrow L_{t}^{adv}-\eta\cdot\nabla J(L_{t}^{adv}, b)$
        \ENDFOR
    \ENDFOR
    \STATE $\gamma=\frac{Q}{||L^{adv}_{t+1}-G_t||}$
    \STATE $L^{adv}_{t+1} =\gamma (L^{adv}_{t+1}-G_t)+G_t$
    \RETURN $L^{adv}_{t+1}$
\end{algorithmic}
\end{algorithm}

\section{Experiments}
In this section, we first describe three benchmark datasets and model architectures used in the experiments. Next, we present our evaluation of the proposed method in the full knowledge setting and the partial knowledge setting, followed by a discussion of the experimental results. We use Tensorflow \cite{tensorflow} to implement ADA and the other baselines. The code is publicly available \footnote{https://github.com/yuweisunn/ADA}.

\subsection{Dataset}
We employed three image classification tasks for conducting the experiments, i.e., MNIST, Fashion-MNIST, and CIFAR-10. These datasets pose different degrees of difficulty for perturbing the FL system. MNIST \cite{mnist} is a handwritten digit image dataset containing 50,000 gray-scale training samples labeled as 0-9 and 10,000 test samples. The size of the images is $28\times28$. Fashion-MNIST \cite{fashion} is an image collection of 10 types of clothing containing 50,000 gray-scale training samples labeled as shoes, t-shirts, dresses, and so on and 10,000 test samples with a size of $28\times28$. CIFAR-10 \cite{cifar10} is a collection of 10 types of objects’ color images, covering 50,000 color training samples labeled as airplane, automobile, and so on and 10,000 test samples. The size of the images is $32\times32\times3$. 

\subsection{Model Architecture}
We applied a four-layer convolutional neural network (CNN) with architecture as follows: the first convolutional layer of the CNN model has a convolution kernel of size 5×5 with a stride of 1 and it takes one input plane and it produces 20 output planes, followed by a ReLU activation function; the second convolutional layer takes 20 input planes and produces 50 output planes and it has a convolution kernel of size 5×5 with a stride of 1, followed by ReLU; the output is flattened followed by a linear transformation of a fully connected layer, which takes as input the tensor and outputs a tensor of size 200; the final fully connected layer outputs a tensor of size 10 representing the 10 categories. Moreover, the categorical cross-entropy is employed to compute the loss. In addition, we apply as a learning function the Adam to update the model with a learning rate of 0.001. The model architecture is shared by all clients and the global model.

\subsection{Numerical Results}
\label{experiments}
We considered a 100-client scenario. Each client had a total of 500 samples randomly drawn from the training set of the applied dataset. Every round, FL randomly selected 10 clients to perform model training thus updating the global model based on the FedAvg. For the local model training, we employed an epoch of one and a batch size of 16. The evaluation was conducted every round after the global update based on the hold-out test set in the applied dataset. We mounted the attack after the global model converged showing no further decrease in the validation loss within the latest 10 rounds. Then the adversary performed ADA by measuring the AD from either the extracted latent feature representations in the full knowledge setting or the shared global model parameters using backward-error analysis in the partial knowledge setting. In addition, though the adversary can mount the attack before the global model converges, it will be less effective due to the difficulty of precisely measuring the AD between different classes.

\subsubsection{ADA with Full Knowledge}
When the global model converged and a compromised client was selected, the malicious client extracted the latent feature representations of different classes in FL by inputting local instances into the shared global model. Then, a latent feature vector with a dimension of 200 was obtained from the last hidden layer of the model. After that, these latent representations were employed to compute the AD between the different classes. 

Furthermore, we compared the performance of the ADA with full knowledge (ADA-full) with the label flipping attack (LF) and the Train and Scale method (TS). For simplicity, we assumed the attacker selected the third label from each applied dataset as the source class. For example, the selected source class is the digit '2' in MNIST. For LF and TS, we computed the average accuracy scores when applying different target classes of the classifier. For ADA, the attacker measured the AD scores between different classes and selected the class with the minimum AD to the source class as the target \ref{eq:adafull}. ATA (ts-ATA) and MTA were evaluated every round with the global model using the hold-out test set of the applied dataset.

\begin{table*}[t]
\centering
  \caption{ATA and MTA with Various Attacking Frequencies}
\begin{tabular}{|p{2.5cm}||p{1cm}|p{1cm}|p{1.2cm}||p{1cm}|p{1cm}|p{1.2cm}||p{1cm}|p{1cm}|p{1.2cm}|}
 \hline
 Attacking Frequency & \multicolumn{3}{|c||}{$\epsilon$=0.01} & \multicolumn{3}{|c||}{$\epsilon$=0.05} & \multicolumn{3}{|c|}{$\epsilon$=0.1}\\ 
 \hline
 Method   & LF    & TS & ADA-full & LF & TS & ADA-full & LF & TS & ADA-full\\
 \hline
  Dataset & \multicolumn{9}{|c|}{ATA}\\
 \hline
 MNIST         & 0.081 & 0.051 & \textbf{0.387} & 0.569 & 0.191 & \textbf{0.832} & 0.912 & 0.678 & \textbf{0.968}\\
 Fashion-MNIST & 0.148 & 0.211 & \textbf{0.695} & 0.364 & 0.548 & \textbf{0.800} & 0.669 & 0.828 & \textbf{0.902}\\
 CIFAR-10  & 0.192 & 0.292 & \textbf{0.463} & 0.321 & 0.400 & \textbf{0.581} & 0.391 & 0.550 & \textbf{0.654}\\
 Average & 0.140 & 0.185 & \textbf{0.515} & 0.418 & 0.380 & \textbf{0.738} & 0.657 & 0.685 & \textbf{0.847}\\
  \hline
 Dataset & \multicolumn{9}{|c|}{MTA}\\
 \hline
 MNIST         & \textbf{0.990} & \textbf{0.990} & \textbf{0.990}	& \textbf{0.990} & \textbf{0.990} & 0.989 & 0.989 & \textbf{0.990}	& 0.989\\
 Fashion-MNIST & 0.809 & 0.810 & \textbf{0.811} & 0.806 & 0.790 & \textbf{0.813} & 0.798 & 0.780 & \textbf{0.810}\\
 CIFAR-10      & 0.590 & 0.596 & \textbf{0.709} & 0.587 & 0.563 & \textbf{0.711} & 0.559 & 0.548 & \textbf{0.702}\\
 Average & 0.796 & 0.799 & \textbf{0.837} & 0.794 & 0.781 & \textbf{0.838} & 0.788 & 0.773 & \textbf{0.834}\\
  \hline 
  \end{tabular}
  \label{table1}
  \end{table*}

\begin{figure}[t]
\centering
\includegraphics[width=0.99\linewidth]{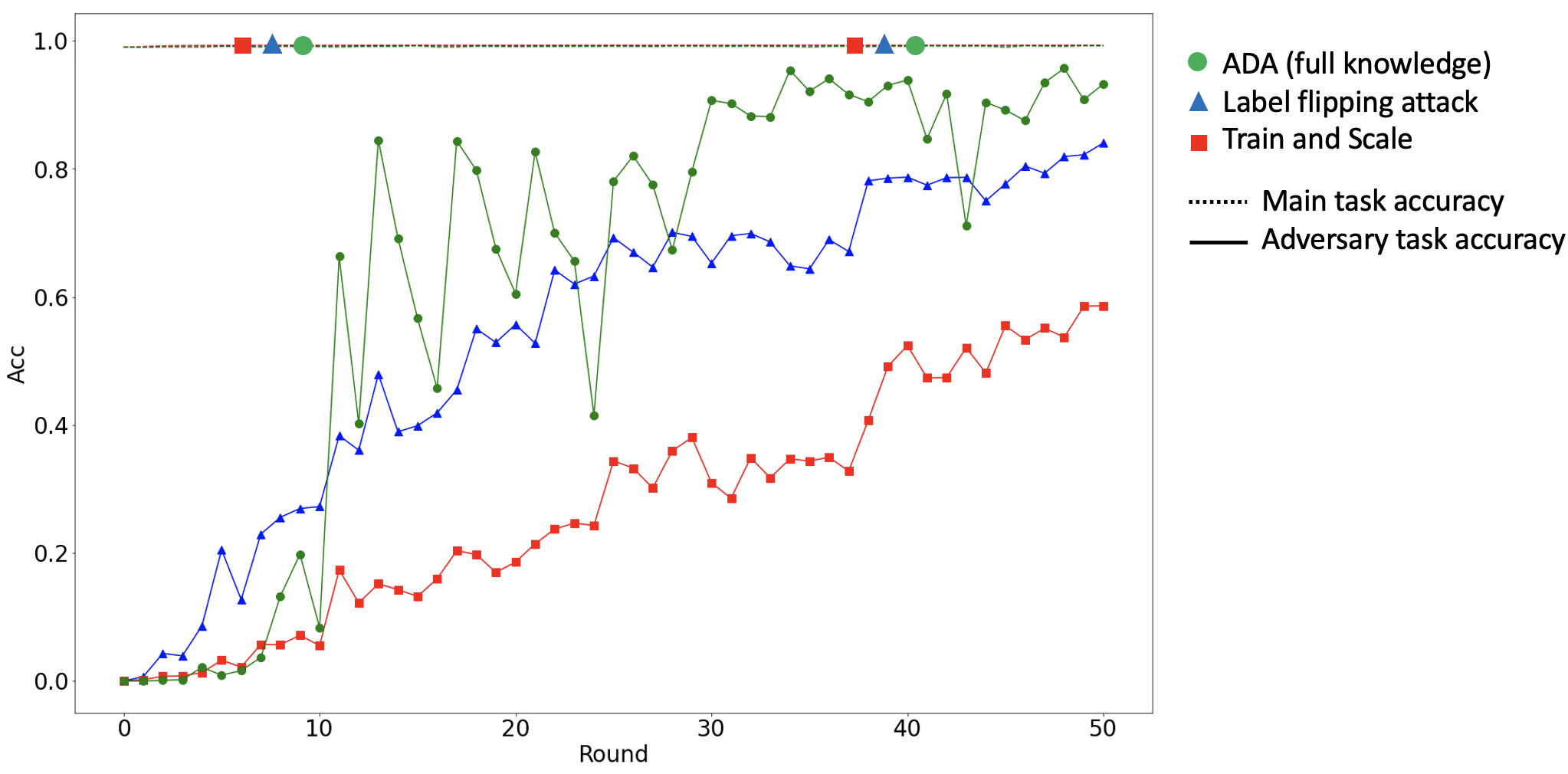}
\caption{Performance comparison among the ADA with full knowledge (ADA-full), the label flipping attack (LF), and the Train and Scale method (TS).}
\label{fig4}
\end{figure}

We first considered there were 10 compromised clients and evaluated the performance of the aforementioned three methods when tackling the MNIST image classification task. Fig. \ref{fig4} illustrates the performance comparison among the three different methods. We further studied the influence of the attacking frequency on the successability of these attacking methods by varying the ratio of compromised clients $\epsilon = \{0.01,0.05,0.1\}$. Table. \ref{table1} shows the numerical results with various attacking frequencies when applying the three different datasets respectively. The final accuracy scores were measured by the maximum values within 50 rounds of FL. The results show that ADA outperforms the other methods with improved ATA and MTA scores when applying various attacking frequencies in all three classification tasks. ADA could achieve an ATA score of 0.387 in MNIST compared with the typical LF method with a score of 0.081 when the attacking frequency was 0.01 where a compromised client was selected around every 10 rounds. Moreover, the improvement of MTA in CIFAR-10 is larger than MNIST and Fashion-MNIST. Finally, we have shown that ADA could improve a poisoning attack's performance in various cases of FL and reduce its influence on the main classification task. 

\begin{figure}[t]
   \begin{subfigure}{0.49\textwidth}
     \centering
    \includegraphics[width=0.8\linewidth]{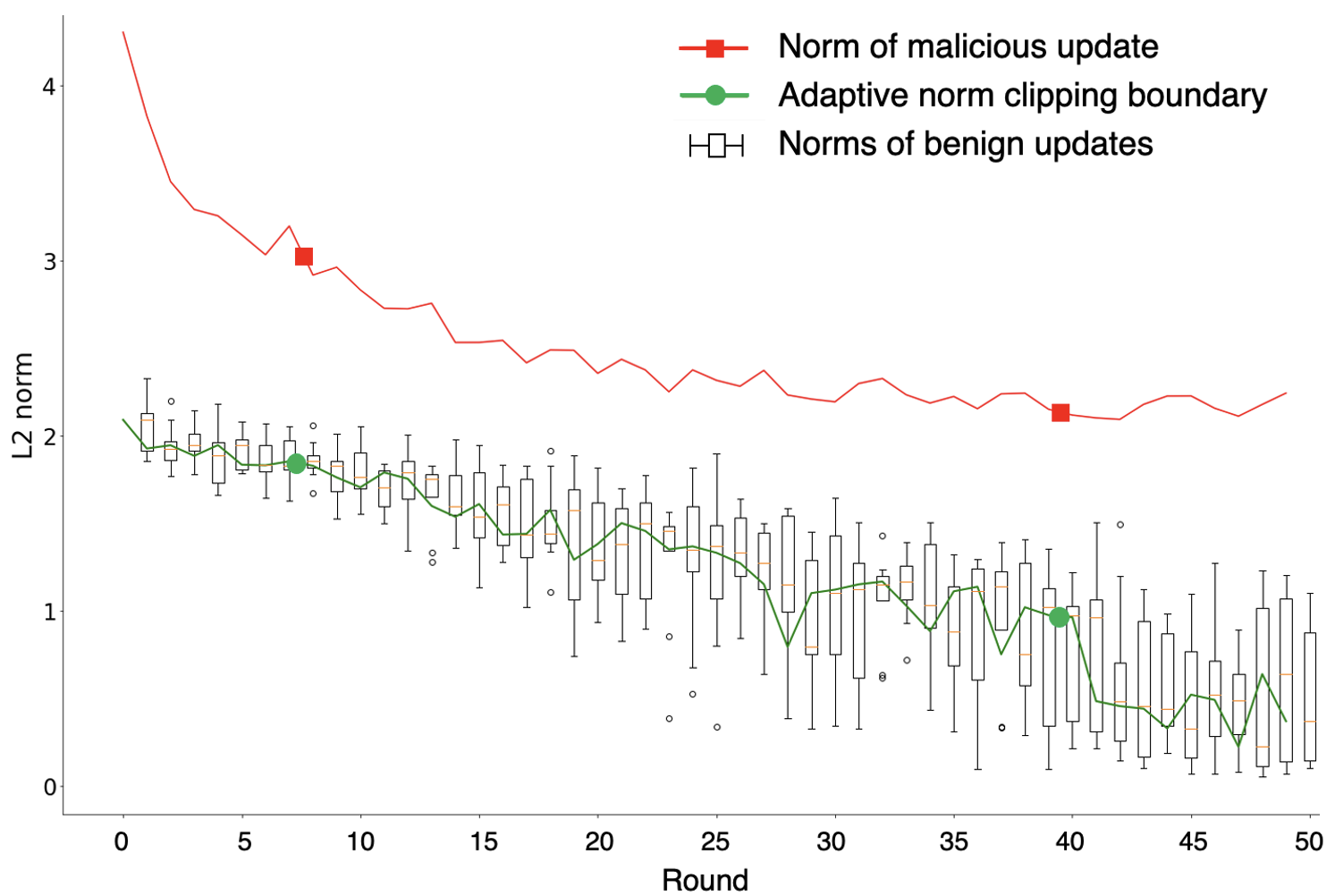}
    \caption{Without the gradient scale adjustment.}
       \label{norm}
   \end{subfigure}\hfill
   \begin{subfigure}{0.49\textwidth}
     \centering
     \includegraphics[width=0.84\linewidth]{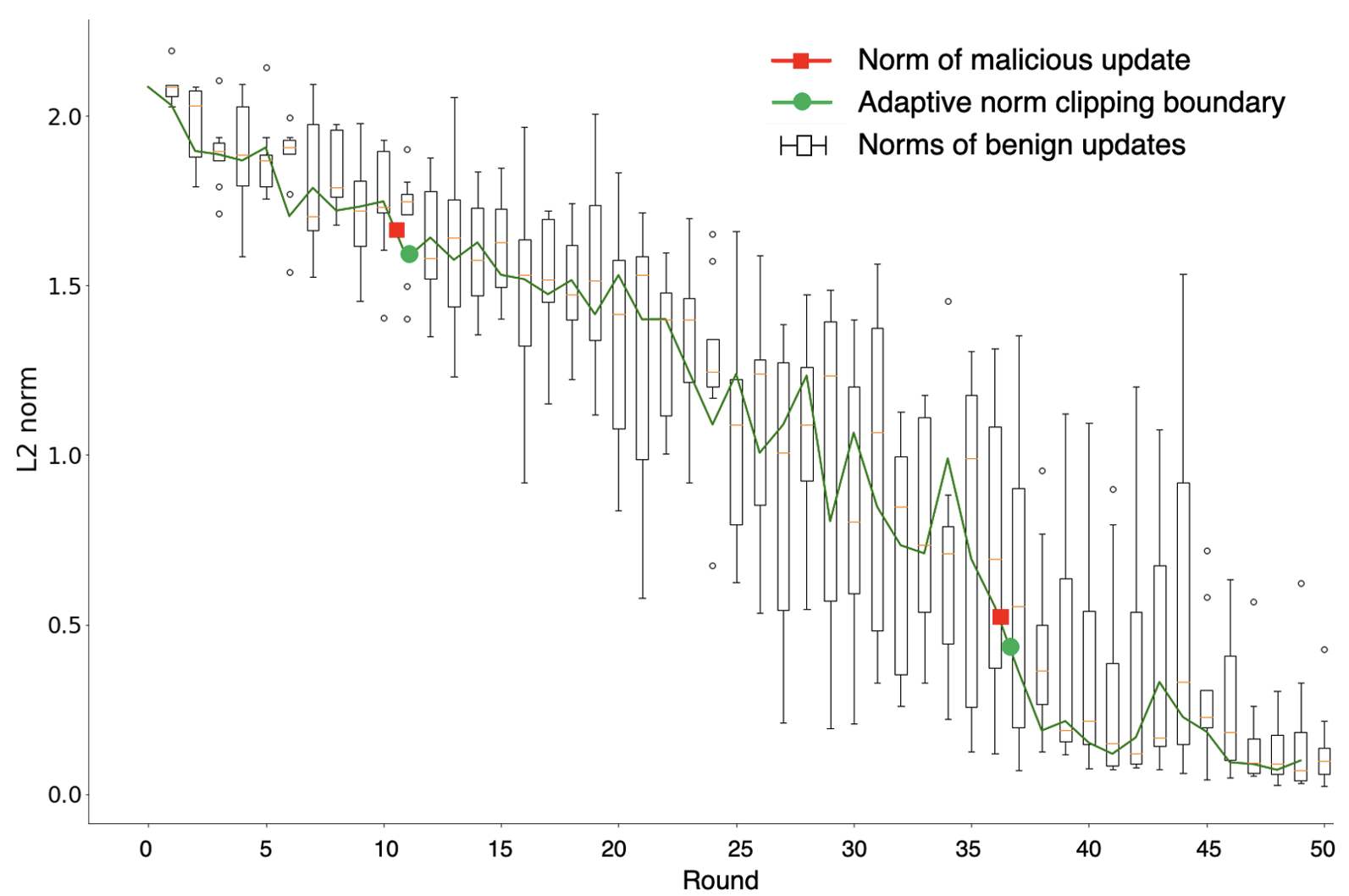}
     \caption{With the gradient scale adjustment.}
        \label{norm2}
   \end{subfigure}
   \caption{Norms of local updates from the benign and malicious clients at each round of FL.}
\end{figure}

In addition, Fig.\ref{norm} illustrates the norms of benign local updates $L^{(k)}_{t+1}-G_t$ and malicious local updates $L^{adv}_{t+1}-G_t$ when applying the typical label flipping attack. Fig.\ref{norm2} illustrates the norms of the benign updates and the scaled malicious updates $\gamma (L^{adv}_{t+1}-G_t)$ using the "Train and Scale" strategy. The $Q$ is estimated by the average of all benign updates' norms.   

\subsubsection{ADA with Partial Knowledge}

The experiments above are based on the assumption that the adversary has full knowledge of the classification task's data distribution. On the contrary, in the partial knowledge setting, we assumed that the entire distribution of a compromised client's local data is unknown to the adversary. The adversary only has access to the source class samples for mounting the attack. To measure the performance of ADA with partial knowledge, we performed the experiments using the same training hyper-parameters and setting. Instead of measuring the AD from the latent feature space, we leveraged the FLAME \ref{eq:flame} to mount the attack. Table. \ref{table2} shows the attack performance in the partial knowledge setting when applying different attacking frequencies with various classification tasks. The results show that ADA can also achieve competing performance in the partial knowledge setting. In addition, in the Fashion-MNIST task, ADA results in degraded performance when $\epsilon ={0.01,0.1}$. This could be due to a more challenging task to deduce the latent feature distribution in Fashion-MNIST. 

\begin{table}[!th]
\centering
  \caption{Evaluation Results of ADA in the Partial Knowledge Setting}
\begin{tabular}{|p{2cm}||p{0.6cm}|p{0.6cm}||p{0.6cm}|p{0.6cm}||p{0.6cm}|p{0.6cm}|}
 \hline
 Attacking Frequency & \multicolumn{2}{|c||}{$\epsilon$=0.01} & \multicolumn{2}{|c||}{$\epsilon$=0.05} & \multicolumn{2}{|c|}{$\epsilon$=0.1}\\ 
 \hline
 Dataset & ATA & MTA & ATA & MTA & ATA & MTA\\
 \hline
 MNIST       &  0.553 & 0.965 & 0.996 & 0.988 &  0.988 & 0.988\\
 Fashion-MNIST &  0.037 &  0.784 & 	0.900 & 0.763 & 0.772 & 0.768\\
 CIFAR-10  & 0.494 & 0.590 & 0.655 & 0.710 & 0.705 & 0.702\\
 Average   & 0.361 & 0.780 & 0.850 & 0.820 & 0.822 & 0.819\\
  \hline 
  \end{tabular}
  \label{table2}
  \end{table}

\section{Conclusion}
We proposed a semi-targeted model poisoning attack on federated learning (FL) called Attacking Distance-aware Attack (ADA). ADA aims to find the optimized target class by measuring the distance of latent feature representation between the source class and the target class. Moreover, in the more challenging partial knowledge setting, FLAME is leveraged to perform backward error analysis on the shared global model thus deducing attacking distances between different classes. Furthermore, we evaluated the proposed method's performance against the metrics of ATA and MTA, with various attacking frequencies and classification tasks. The result showed that the semi-targeted ADA could increase the attack performance while retaining legitimate tasks' performance in the various cases of FL. The goal is to introduce this new type of semi-targeted poisoning attack and reveal the risk. In future, a generator model that produces adversarial samples based on the revealed attacking distance information can be adopted to mount semi-targeted backdoors \cite{29} in FL. In addition, other defense strategies against the poisoning attacks in FL besides the norm-based methods will be further studied, such as the Byzantine-robust aggregation \cite{BlanchardMGS17,MhamdiGR18}.

\section*{Acknowledgment}
This paper was partially supported by the JRA program at RIKEN AIP and JSPS KAKENHI Grant Number JP22J12681. We thank all the anonymous reviewers for their constructive comments and suggestions.

\bibliographystyle{IEEEtran}
\bibliography{references}

\end{document}